\documentclass[letterpaper, 10 pt, conference]{ieeeconf}  

\IEEEoverridecommandlockouts  

\usepackage{amssymb}
\usepackage{graphicx}
\usepackage{xcolor}
\usepackage{subcaption}
\usepackage{amsmath}
\usepackage{fontawesome5}
\usepackage[colorlinks=true, linkcolor=blue, citecolor=blue, urlcolor=blue]{hyperref}

\title{\LARGE \bf

He3-Seeker: Robotic Information Planning for Lunar Helium-3 Distribution Mapping
}
\author{
Dong Li$^{1,8,*}$, Yujie Zheng$^{2,*}$, Chengdeng Cao$^{3,8}$, Siyu Teng$^{4}$, Yuchen Li$^{5,8}$, Yang Gao$^{6}$, Long Chen$^{7,8,\ddagger}$
\thanks{* Dong Li and Yujie Zheng contributed equally to this work.}
\thanks{$\ddagger$ Corresponding Author: Long Chen (long.chen@ia.ac.cn)}
\thanks{$^{1}$ Dong Li is with the Institute of Automation, Chinese Academy of Sciences, China, and also with the Faculty of Innovation Engineering, Macau University of Science and Technology, Macau (e-mail: doongli@ieee.org)}%
\thanks{$^{2}$ Yujie Zheng is with the College of Artificial Intelligence, China University of Petroleum (Beijing) (e-mail: 2024011639@student.cup.edu.cn)}%
\thanks{$^{3}$ Chengdeng Cao is with the School of Earth and Space Science and Technology, Wuhan University, China (e-mail: caochengdeng@163.com)}%
\thanks{$^{4}$ Siyu Teng is with the College of Civil and Transportation Engineering, Shenzhen University, China (e-mail: siyuteng@szu.edu.cn)}%
\thanks{$^{5}$ Yuchen Li is with the Department of Informatics of the Technical University of Munich, Germany, and also with the University Research and Innovation Center, Obuda University, Hungary (e-mail: yuchen.li@tum.de)}%
\thanks{$^{6}$ Yang Gao is with the Centre for AI Robotics in Space Sustainability, Department of Mechanical and Aerospace Engineering, Hong Kong University of Science and Technology, Hong Kong (e-mail: yanggao@ust.hk)} %
\thanks{$^{7}$ Long Chen is with the Institute of Automation, Chinese Academy of Sciences, China, and also with the School of Mechanical and Electrical Engineering, China University of Mining and Technology Beijing, China, and also with the University Research and Innovation Center, Obuda University, Hungary}
\thanks{$^{8}$ Dong Li, Chengdeng Cao, Yuchen Li and Long Chen are with OpenSpace Lab, China} %
}

\begin{document}

\maketitle
\thispagestyle{empty}
\pagestyle{empty}

\begin{abstract}

Lunar helium-3 is a highly valuable strategic resource, pivotal to the advancement of both deep-space exploration and space mining.
Existing lunar helium-3 exploration methodologies rely primarily on indirect measurements via remote sensing, which are often characterized by limited precision, low reliability, and insufficient spatial resolution.
In this paper, we introduce He3-Seeker, an active robotic exploration method for helium-3 distribution mapping. First, we provide a formal definition of the active helium-3 exploration problem. We then develop the He3-Seeker framework, which is conceptually based on multi-point drilling, sampling, and in situ analysis. In particular, we use robotic information planning to guide autonomous robot navigation and active sensing.
Additionally, to evaluate the proposed algorithm, we introduce a reliable method for generating reference data of lunar helium-3 distribution based on low-resolution orbital remote sensing measurements.
Simulation experiments verify that He3-Seeker enables fast, high-fidelity helium-3 distribution for resource exploration tasks.
Code and simulation environments are available at \href{https://github.com/OpenSpace-Lab/He3-Seeker}{\color{blue}https://github.com/OpenSpace-Lab/He3-Seeker}.
\end{abstract}

\section{INTRODUCTION}
Space resource utilization serves as a critical enabler and core methodology for sustaining long-term space exploration and establishing permanent extraterrestrial habitats~\cite{li2026mining}. Given its proximity to Earth, the Moon has witnessed an unprecedented surge in robotic and autonomous exploration initiatives over recent years. In particular, flagship endeavors such as China's Chang’E Lunar Exploration Program and NASA's Artemis Program have both designated autonomous lunar mining and resource extraction as pivotal strategic objectives~\cite{li2019china, smith2020artemis}. 
Among these target resources, helium-3 represents an exceptionally high-value asset due to its extreme scarcity on Earth. As a clean energy source, helium-3 holds profound potential as a primary fuel for controlled nuclear fusion and next-generation rocket propellants, making its detection and extraction a paramount focus of modern lunar exploration~\cite{fa2007quantitative,simko2014lunar,ding2023moon}.

\begin{figure}[t!]
    \centering
    \includegraphics[width=0.47\textwidth]{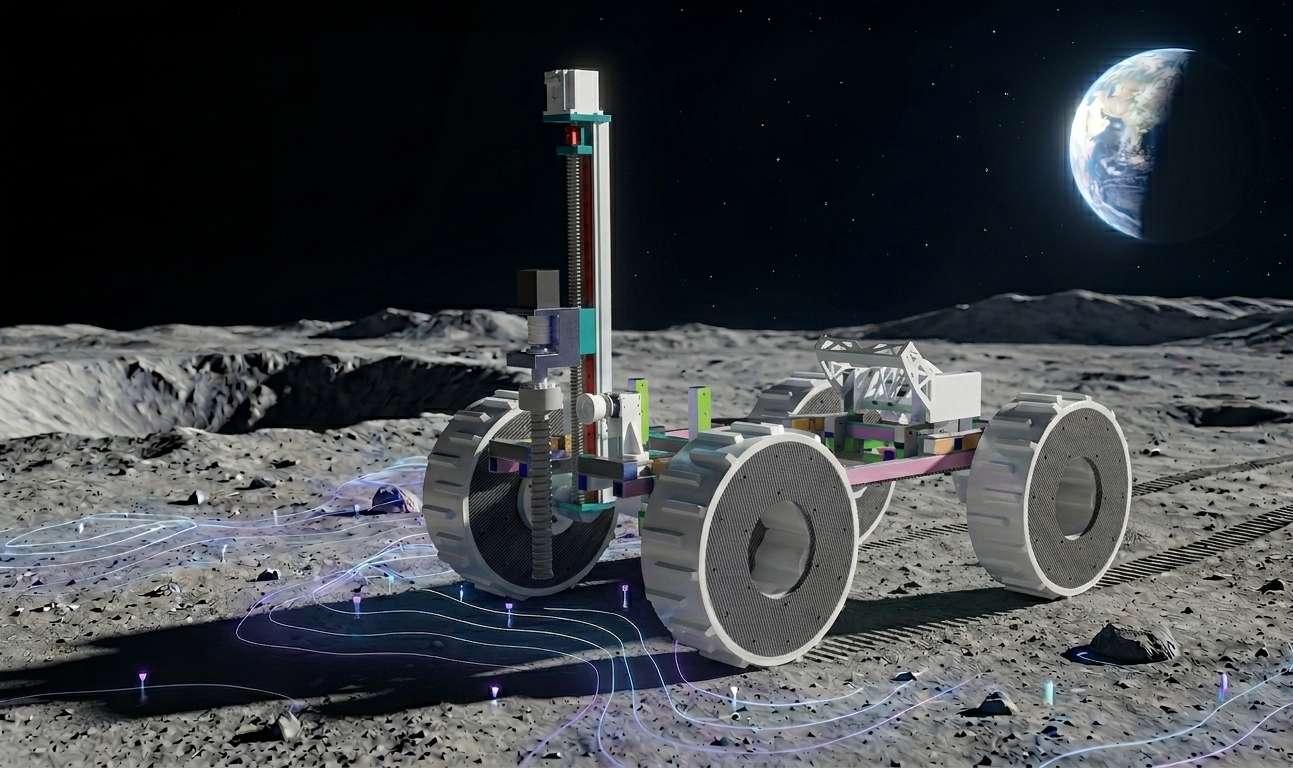} 
    \caption{\textbf{Schematic Diagram of Robotic Lunar Helium-3 Distribution Exploration.} The exploration rover, equipped with an integrated drilling-and-sampling payload alongside an onboard compositional analysis system, addresses a critical question: how to dynamically map the \textbf{high-precision dense spatial distribution} of helium-3 within the target region within a significantly \textbf{shortened timeframe}? \textit{(Figure rendered with assistance from Nano Banana.)}}
    \label{fig:he3_intro} 
\end{figure}

From the perspective of the entire resource extraction workflow, initial exploration represents the foundational and most critical phase. Currently, lunar helium-3  exploration relies heavily on indirect orbital remote sensing. Since the Moon lacks a thick atmosphere, helium-3  ions delivered by the solar wind are directly implanted into the lunar regolith~\cite{geiss2004apollo}. Sample analyses from the Apollo and Chang'E missions revealed that helium-3 is primarily retained within ilmenite due to its superior radiation resistance, existing in either a loosely bound pore state or a permanently trapped lattice-implanted state influenced by particle size effects~\cite{jolliff2000major,zhang2025he}. Hence, mainstream exploration frameworks utilize gamma-ray spectroscopy, optical imaging, and ground-penetrating radar to map ilmenite abundance, thereby inferring helium-3 spatial distributions~\cite{ding2023moon}. However, this indirect estimation paradigm suffers from fundamentally low spatial resolution, compromised measurement precision, and poor reliability under complex terrain constraints.

To address these limitations and achieve high-fidelity resource mapping, several surface rover missions are currently being planned by space agencies in China and the US, as well as commercial enterprises such as Interlune. Specifically, NASA's VIPER rover, targeted for deployment to the lunar polar regions, demonstrates the viability of integrating drilling mechanisms with mass spectrometers to detect volatile species and water ice~\cite{smith2022viper,elphic2015simulated}. However, extending this paradigm to helium-3 presents a critical, unresolved challenge: how to orchestrate multi-point drilling and localized isotopic analysis so that a mobile rover can reconstruct an accurate resource distribution map within a minimized timeframe. Current planetary robotics literature still lacks a systematic methodology to balance this trade-off between the temporal cost of multi-point drilling and the resulting mapping accuracy.

In this paper, we focus on a lunar rover equipped with drilling, sampling, and in-situ analysis capabilities. Our primary objective is to develop an algorithmic framework that enables high-accuracy mapping of lunar helium-3 distribution within a minimized operational timeframe. The main contributions of this paper are highlighted as follows.

First, we formally define the lunar helium-3 resource spatial exploration and mapping problem for a multi-functional lunar rover under practical operational constraints, such as coupled drilling-sampling-analysis sequences, and establish a well-approximated reference model for lunar helium-3 distribution to serve as an evaluation benchmark.

Second, we develop \textbf{He3-Seeker}, a data-efficient autonomous exploration framework that uses robotic information planning (RIP) to guide autonomous robot navigation and active sensing. By exploiting RIP to systematically maximize information gain along the robot's trajectory, the framework drastically accelerates the exploration process while ensuring high-fidelity resource distribution mapping.

Third, we construct a simulation environment through extensive comparative analyses against baselines to validate the efficacy of our approach. To benefit the space robotics community, both our source code and evaluation benchmarks will be made publicly available upon acceptance of the paper.

A conceptual overview of the lunar helium-3 spatial distribution exploration is depicted in Fig.~\ref{fig:he3_intro}.

\begin{figure*}[t!]
    \centering
    \includegraphics[width=1\textwidth]{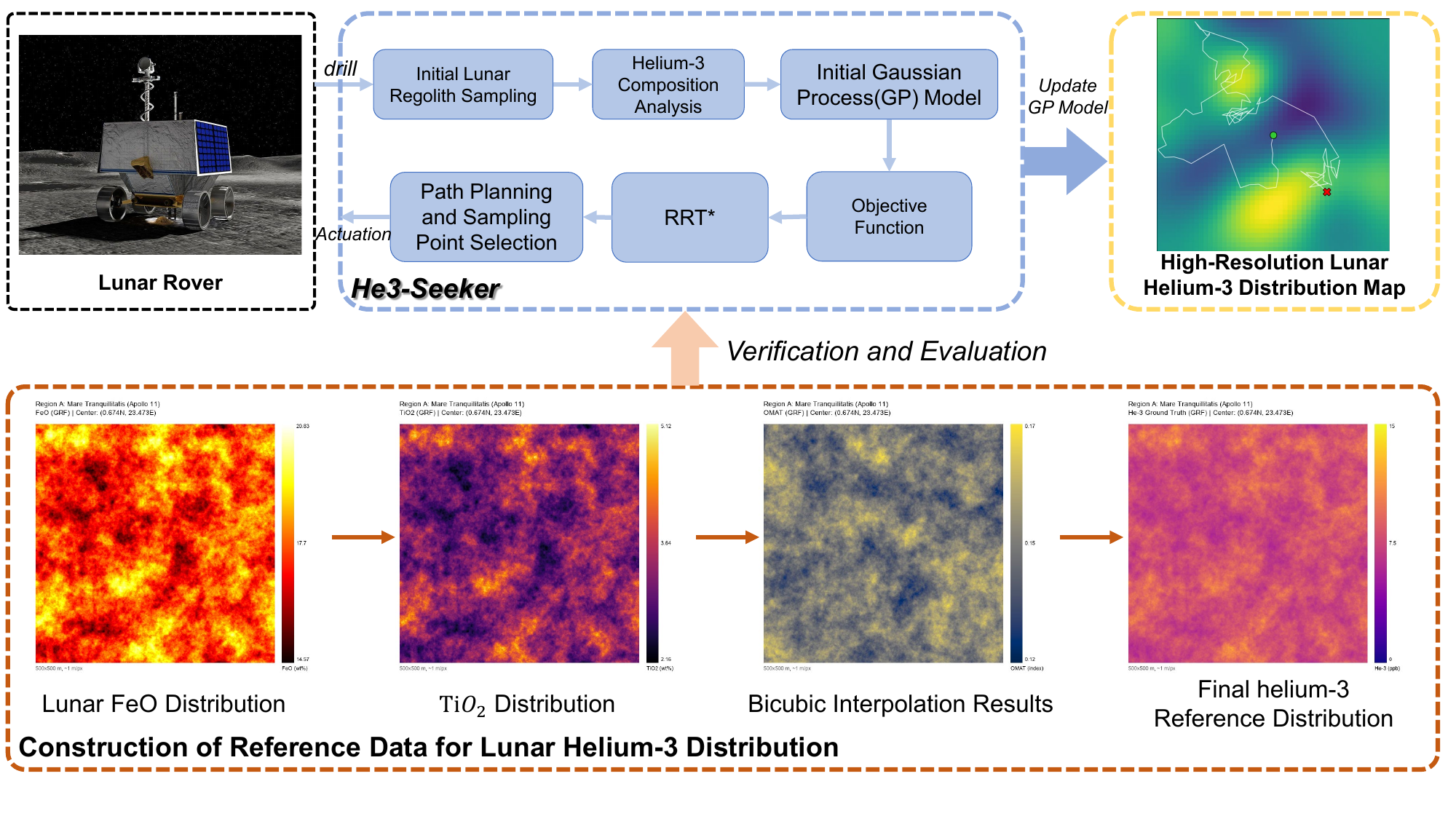} 
    \vspace{-20pt}
    \caption{\textbf{Overview of the He3-Seeker Framework}. (Top) The systemic architecture of \textit{He3-Seeker}, which guides a rover equipped with single-point helium-3 measurement capability to navigate and conduct multi-point sampling for dense lunar helium-3 distribution mapping. (Bottom) The generation of a dense lunar helium-3 reference map utilized for algorithmic evaluation and testing.}
    \label{fig:pipeline} 
\end{figure*}

\section{Related Works}

\subsection{Lunar Helium-3 Resource Prospecting}
Resource exploration serves as the prerequisite for extraterrestrial mining. Currently, the exploration of lunar helium-3 heavily relies on remote sensing coupled with rover-based in-situ detection. At the macro scale, the Chang’E missions and the Lunar Reconnaissance Orbiter (LRO) utilize visible-to-infrared reflection spectrometers to invert helium-3 abundance~\cite{lucey2000lunar}, exploiting the distinctive absorption spectra of ilmenite, a primary carrier of helium-3, at specific wavelengths. Concurrently, multi-frequency microwave radiometers leverage their penetrative imaging capabilities to estimate macroscopic three-dimensional reserves via regolith thickness inversion models~\cite{fa2007quantitative}, while neutron and gamma-ray spectrometers quantitatively analyze key elemental abundances based on energy spectrum signatures of secondary radiation excited by cosmic rays~\cite{lawrence1998global}. At the micro and engineering scales, Ground Penetrating Radar (GPR) is widely deployed, achieving fine subsurface stratigraphic profiling and discrete rock identification by exploiting the reflection mechanism of high-frequency electromagnetic waves at discontinuous interfaces within heterogeneous lunar regolith. Consequently, this yields high-resolution geometric and physical priors to facilitate rover path planning and mineral identification~\cite{ding2023moon, fang2014lunar}.

However, such indirect remote sensing approaches lack fine spatial resolution and suffer from substantial estimation uncertainties during data inversion. To address these limitations, advanced lunar rovers have been developed for high-precision, localized resource mapping. For instance, NASA’s RESOLVE, a rover-borne payload equipped with helium-3 gas proportional counters, enables real-time detection of hydrogen and volatile species associated with lunar exploration~\cite{elphic2015simulated}. Furthermore, NASA’s VIPER rover advances this paradigm by integrating deep regolith drilling and in-situ analysis capabilities, aiming to construct high-fidelity resource distribution maps across the lunar polar regions~\cite{smith2022viper}.
However, how to guide the robot to construct highly accurate resource distribution maps while minimizing both the traversal distance and the number of discrete sampling sites remains an open and unaddressed challenge.

\subsection{Active Exploration and Information Gathering}
To map unknown spatial fields, resource prospecting is commonly formulated as an active exploration task. Within this context, robotic information gathering (RIG) frameworks are widely employed to maximize the information gain regarding the under-sampled field under vehicle kinematic and energy constraints. Generally, the core objective of RIG is formalized as finding an optimal trajectory that maximizes an information-theoretic metric (e.g., mutual information or entropy reduction) over a spatial field modeled by a probabilistic estimator, typically a Gaussian Process.

Recent advancements have significantly enhanced the efficiency and accuracy of RIG algorithms. For instance, AK~\cite{chen2022ak} introduced adaptive kernel configurations to capture non-stationary spatial correlations, while the Probabilistic Online Attentive Mapping (POAM) framework~\cite{chen2024poam} leveraged real-time belief updates to guide exploration in dynamic or highly uncertain environments. 
~\cite{qiao2026multi} developed an intent-aware cooperative planning method that maintains Gaussian Process (GP)-based interest and risk beliefs to balance information gain and operational safety.
To exploit coupling relationships between cross-modality signals, coregionalized GP formulations integrated with spatiotemporal kernels~\cite{booth2023informative} were developed to jointly model multiple correlated dynamic resource attributes, thereby accelerating field convergence. From a path generation perspective, sample-based heuristics such as integrating Rapidly-exploring Random Trees with the online learning of GPs (RRT-GP)~\cite{viseras2019robotic} have been utilized to efficiently explore informative trajectories and reconstruct spatial fields across complex geometric terrains.

Despite these algorithmic insights, a critical open challenge remains: how to tailor these active exploration strategies to the unique enrichment characteristics of lunar helium-3, while simultaneously guaranteeing safe and efficient navigation over non-flat, hazardous lunar terrains under extremely sparse discrete sampling.

\section{Problem Formulation}

\subsection{Objective and Task Definition} \label{sec:define}

\subsubsection{Problem Scenario and Prior Knowledge}
First, it is assumed that helium-3 is known to exist within a target spatial region $\mathcal{X} \subset \mathbb{R}^2$. This prior knowledge typically originates from rough observations with relatively low accuracy, such as satellite remote sensing, which merely confirms the presence of helium-3 without providing detailed metrics. The true underlying density distribution of helium-3 across the region is modeled as an unknown continuous field $f(x): \mathcal{X} \rightarrow \mathbb{R}^+$.

\subsubsection{Robotic Sampling and Observation Model}
As illustrated in Fig. 1, the exploration rover features an integrated drilling-sampling payload and an onboard compositional analysis system. When the rover halts at a discrete spatial coordinate $x_t \in \mathcal{X}$ at time step $t$, it performs a localized drilling and isotope analysis operation. The discrete observation obtained by the rover can be formulated as:

\begin{equation}
y_t = f(x_t) + \epsilon_t, \quad \epsilon_t \sim \mathcal{N}(0, \sigma_n^2)
\end{equation}
where $y_t \in \mathbb{R}$ represents the measured  helium-3 concentration value, and $\epsilon_t$ is the independent and identically distributed measurement noise with variance $\sigma_n^2$. 

After $t$ sampling steps, the rover accumulates a sparse, discrete historical dataset $\mathcal{D}_t = \{(x_i, y_i)\}_{i=1}^t$. The dense distribution mapping task is derived through a regression framework (e.g., Gaussian Process Regression), formulating the posterior distribution:
\begin{equation}
\hat{f}(x) \sim \mathcal{GP}\left(\mu_t(x), \Sigma_t(x, x')\right)
\end{equation}
where $\mu_t(x)$ represents the predicted high-precision dense mean distribution, and $\Sigma_t(x, x')$ quantifies the epistemic uncertainty under sparse sampling.

\subsubsection{Active Exploration Task Formulation}
Despite the theoretical feasibility of field reconstruction, the active exploration strategy must resolve a multi-objective optimization problem under strict physical constraints. Let $\tau = \{x_1, x_2, \dots, x_N\}$ define the sequential trajectory of the rover. The core challenge is to optimize the next optimal sampling location $x_{t+1}$ and the collision-free path $\gamma_t: x_t \rightarrow x_{t+1}$. 
In summary, the task mandates planning a collision-free path $\gamma$ along with a sequence of discrete sampling locations $\{x_1, x_2, \dots, x_N\} \in \tau$, enabling the high-precision estimation of the dense lunar helium-3 field $\hat{f}(x)$ within a shorter timeframe.

\subsection{Construction of  Reference Data for Helium-3 Distribution} \label{sec:he3}

Given the current lack of high-resolution lunar helium-3 distribution maps suitable for robotic exploration, we establish a high-resolution helium-3 reference data to serve as a benchmark for evaluating subsequent active robotic sensing missions.
Specifically, the underlying helium-3 distribution is derived through a chemical back-calculation pipeline based on iron oxide ($\text{FeO}$) and titanium dioxide ($\text{TiO}_2$) abundances. 
To synthesize the benchmark $500\,\mathrm{m} \times 500\,\mathrm{m}$ of lunar helium-3 distribution, we implement a five-step physically consistent simulation pipeline based on planetary remote sensing data:

\begin{enumerate}
    \item \textbf{Data Acquisition:} The baseline iron oxide ($\text{FeO}$) and optical maturity ($\text{OMAT}$) abundance maps are sourced from the Kaguya Multiband Imager (MI)~\cite{lemelin2019compositions} mosaics provided by the USGS Astrogeology Science Center\footnote{\url{https://astrogeology.usgs.gov/search/map/lunar-kaguya-multiband-imager-mosaics}}.
    
    \item \textbf{Geochemical Conversion ($\text{FeO} \rightarrow \text{TiO}_2$):} The baseline titanium dioxide ($\text{TiO}_2$) abundance is derived using the established empirical relationship \cite{lucey1998mapping}:
    \begin{equation}
    \text{TiO}_2 = 10^{0.06 \times \text{FeO} - 0.54} \quad (\text{wt}\%)
    \end{equation}
    
    \item \textbf{Bicubic Interpolation and Sub-Pixel Downsampling:} The original satellite data ($\sim 59\,\text{m/pixel}$) is resampled onto a $500 \times 500$ fine grid covering a $500 \times 500\,\text{m}$ target region using bicubic interpolation. This ensures that the macro-scale spatial gradients are governed entirely by real remote sensing observations.
    
    \item \textbf{Joint Gaussian Random Field (GRF) Perturbation:} To model sub-pixel topography and geological variations ($< 59\,\text{m}$) without high-resolution inversion data, a zero-mean Gaussian Random Field (GRF) is introduced to simulate small-scale features (e.g., small craters and ejecta blankets). The perturbations are applied multiplicatively to the raw inputs ($\text{FeO}$ and $\text{OMAT}$) via Cholesky decomposition rather than directly modifying the derivative products:
    \begin{equation}
    X' = X \times (1 + \sigma \cdot \text{GRF})
    \end{equation}
    where the correlation length is set to $\ell = 25\,\text{m}$ and the perturbation amplitude is $\sigma = 5\%$.
    
    \item \textbf{Physical Consistency and helium-3 Estimation:} Crucially, derivative layer values are re-computed using the perturbed raw inputs to preserve strict physical consistency:
    \begin{equation}
    \text{TiO}_2' = 10^{0.06 \times \text{FeO}' - 0.54}
    \end{equation}
    Finally, the high-resolution helium-3 concentration (in ppb) is quantified at each grid point based on the Fa-Jin physical formulation \cite{fa2007quantitative}:
    \begin{equation}
    ^3\text{He} = f(\text{TiO}_2', \text{OMAT}', F)
    \end{equation}
    where $F$ denotes the localized solar wind flux determined by the region's latitude \cite{fa2007quantitative}. This derived field automatically preserves the nonlinear geological correlation between $\text{FeO}$, $\text{TiO}_2$, and $\text{OMAT}$ without requiring heuristic clipping or mean correction.
\end{enumerate}

The visualization effect of the generated helium-3 reference data is shown in the bottom rows of Fig.~\ref{fig:pipeline}.

\section{He3-Seeker Framework}

\subsection{Overview}

We assume that the robot is equipped with lunar soil sampling and compositional analysis capabilities, enabling it to measure the helium-3 concentration value, denoted as $y_t \in \mathbb{R}$.
As stated in Sec.~\ref{sec:define}, the overarching objective is to plan a trajectory $\gamma$ alongside a sequence of discrete sampling locations $\{x_1, x_2, \dots, x_N\} \in \tau$ within the target spatial region $\mathcal{X}$. This planning framework aims to accurately estimate the dense lunar helium-3 field $\hat{f}(x)$ under strict time constraints. The system framework is shown in Fig.~\ref{fig:pipeline}.

We adopt a two-step robotics information gathering framework. The framework consists of: 
(i) a global \textbf{Objective Function} step that leverages Gaussian Process (GP) environment modeling and information gain computation to evaluate highly informative sampling coordinates under a travel budget; and 
(ii) an \textbf{Informative Path Planner} based on $\text{RRT}^*$ to refine the sampling trajectory by balancing the information gain and path cost. 

\subsection{Robotic Information Planning with Gaussian Process}

\subsubsection{Gaussian Process Modeling}
The unknown continuous distribution field of helium-3 is modeled as a spatial Gaussian Process: $f(x) \sim \mathcal{GP}(m(x), k(x, x'))$, where $m(x)=0$ denotes the prior mean and $k(x, x')$ is a Squared Exponential (SE) covariance function:
\begin{equation}
k(x, x') = \sigma_f^2 \exp\left(-\frac{\|x - x'\|^2}{2\ell^2}\right) + \sigma_n^2 \delta_{xx'}
\end{equation}

Given a history of discrete rover drilling measurements $\mathbf{y} = [y_1, \dots, y_n]^T$ at coordinates $\mathbf{X} = [x_1, \dots, x_n]^T$, the posterior distribution at an unvisited probe location $x_*$ is conditioned as a Gaussian $\mathcal{N}(\mu_*, \Sigma_*)$ via Gaussian Process Regression (GPR):
\begin{align}
\mu_* &= \mathbf{K}_*^T \mathbf{K}^{-1} \mathbf{z} \\
\Sigma_* &= k(x_*, x_*) - \mathbf{K}_*^T \mathbf{K}^{-1} \mathbf{K}_*
\end{align}
where $\mathbf{K}$ and $\mathbf{K}_*$ represent the respective training and cross-covariance matrices.

\subsubsection{Information Gain Metric}

To evaluate the potential of candidate sampling paths, we adopt the mean posterior differential entropy as the core information metric, which reflects the reduction of field uncertainty. For a predicted node or path segment $\mathcal{P}$, the localized information gain $I(\mathcal{P})$ is formulated directly using the GP posterior covariance $\Sigma_*$:
\begin{equation}
I(\mathcal{P}) = \frac{1}{2} \log |2\pi e \mathbf{\Sigma}_*(\mathcal{P})|
\end{equation}

Maximizing this metric drives the rover toward regions with sparse measurements and high helium-3 density uncertainty.

\subsection{RRT*-based Informative Path Planner}
To avoid myopic decisions and optimize the exploration trajectory, we adapt an online $\text{RRT}^*$-based~\cite{karaman2011sampling}  informative path planner to systematically guide the rover. At each planning iteration, the planner evaluates potential target locations within the target region $\mathcal{X}_{\text{free}}$ that maximize information gain under a remaining travel budget $b$. This planner incorporates a multi-objective utility function $u(\mathcal{P})$ to trade off information accumulation against path cost:
\begin{equation}
u(\mathcal{P}) = \frac{I(\mathcal{P})^\alpha}{c(\mathcal{P})}
\end{equation}
where $I(\mathcal{P})$ represents the cumulative information gain along path $\mathcal{P}$, $c(\mathcal{P})$ denotes the associated travel cost, and $\alpha$ is a scaling coefficient balancing the trade-off. By executing $\text{RRT}^*$ sampling and tree-rewire operations optimized for this utility, the rover asymptotically converges to an informative, cost-efficient path to conduct subsequent localized drilling and isotope analysis.

\begin{figure}[t!]
    \centering
    \includegraphics[width=0.5\textwidth]{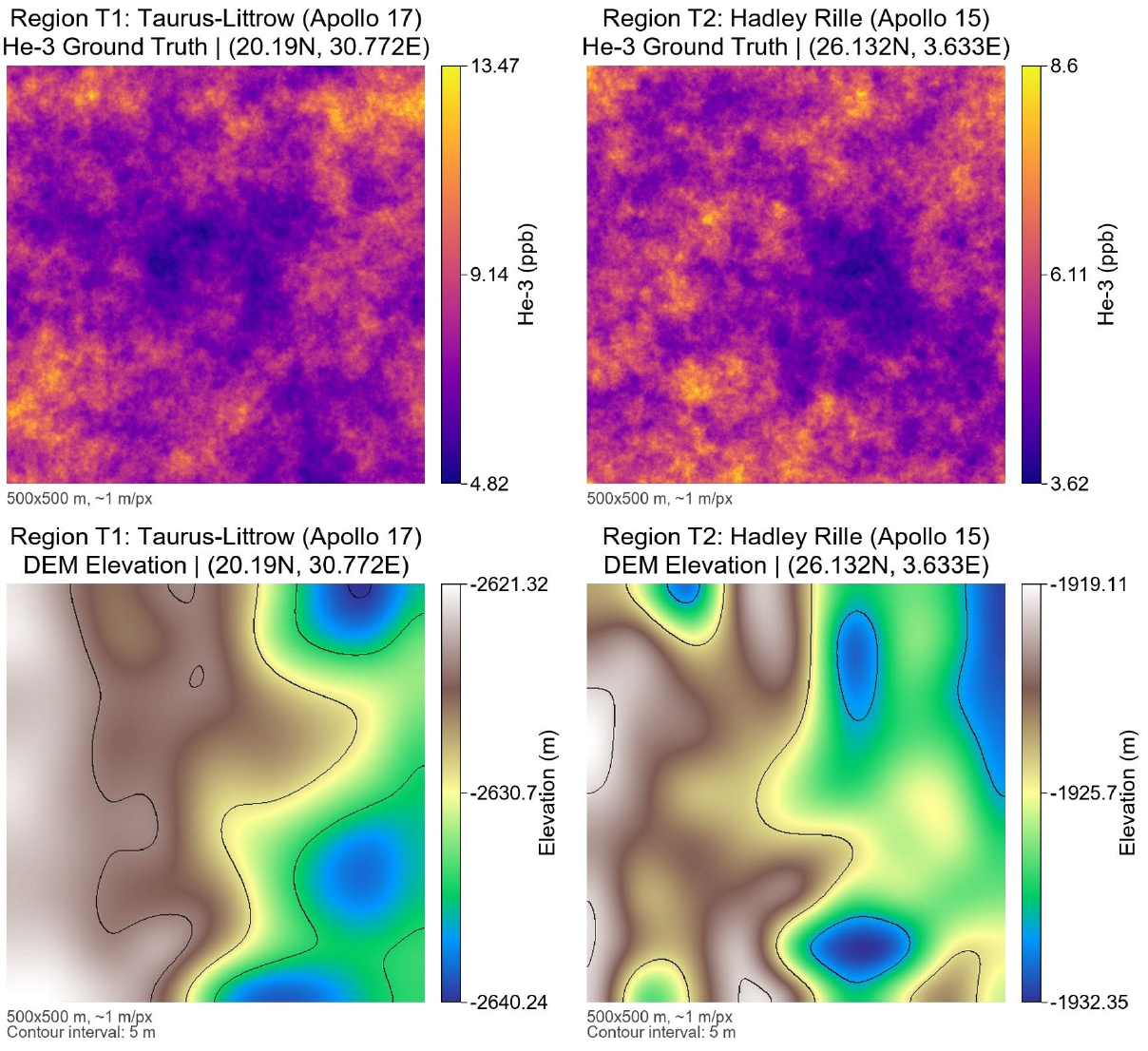} 
    \vspace{-10pt}
    \caption{The reference data of helium-3 distribution and the corresponding digital elevation models (DEM) for trafficability detection in the two constructed regions.}
    \label{fig:scenes} 
\end{figure}

\begin{figure}[htbp]
    \centering
    \begin{subfigure}[b]{0.48\textwidth}
        \centering
        \includegraphics[width=\textwidth]{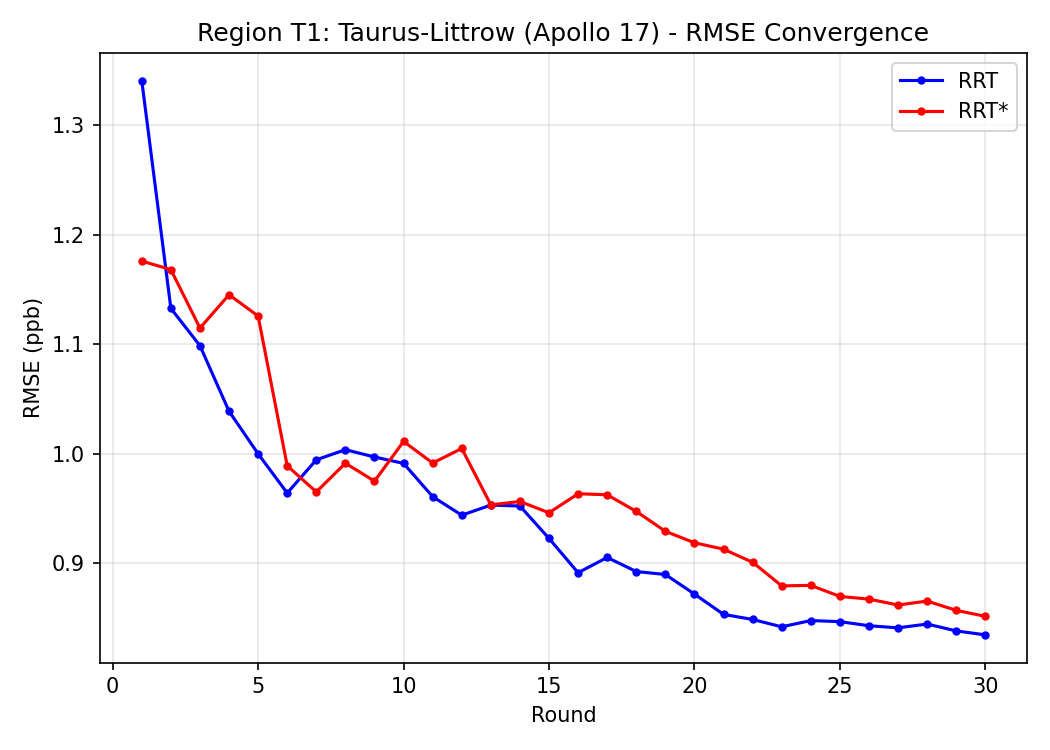} 
            \vspace{-15pt}
        \caption{RMSE in Region T1.}
        \label{fig:sub_framework_a}
    \end{subfigure}
    \hfill
    \begin{subfigure}[b]{0.48\textwidth}
        \centering
        \includegraphics[width=\textwidth]{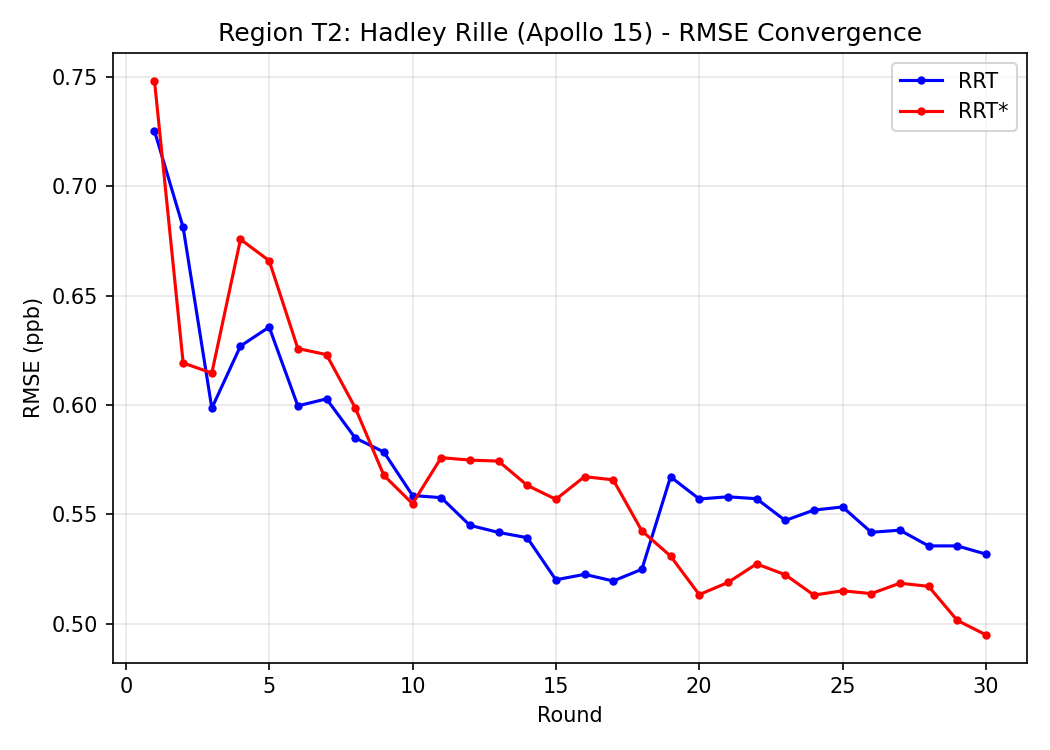}
            \vspace{-15pt}
        \caption{RMSE in Region T2.}
        \label{fig:sub_framework_b}
    \end{subfigure}
    \caption{Comparison between RRT* (ours) and RRT (baseline) in Region T1 (Taurus-Littrow) and Region T2 (Hadley Rille). A round comprises station selection, path planning, and sampling, with each round yielding 5 sampling points (totaling 150 points across 30 rounds).} 
    \label{fig:RMSE}
\end{figure}

\section{Experiments}

\subsection{Simulation Environment Setup}
To comprehensively evaluate the performance of the proposed active information gathering framework, we construct two distinct simulation environments based on actual lunar geological characteristics. 
Following the method described in Sec.~\ref{sec:he3}, we enhance the resolution of the mapping data to a $500 \times 500\,\text{m}$ grid to sufficiently satisfy the requirements for rover-level detailed exploration.

\begin{itemize}
    \item \textbf{T1: Taurus-Littrow (Apollo 17 Landing Site Area)} \\
    Centering at coordinates $(20.190^\circ\text{N}, 30.772^\circ\text{E})$, this region is situated within the Taurus-Littrow valley. Geological data from the Apollo 17 mission confirm that this area exhibits a high variation in solar wind-implanted elements, with the helium-3 concentration ranging from $4.82$ to $13.47\,\text{ppb}$. Region T1 serves as a high-concentration benchmark to evaluate the mapping efficiency under complex geological conditions.
    
    \item \textbf{T2: Hadley Rille (Apollo 15 Landing Site Area)} \\
    Centering at coordinates $(26.132^\circ\text{N}, 3.633^\circ\text{E})$, this region is located near the Hadley–Apennine region. Characterized by its distinctive rille structures and basaltic plains, the helium-3 concentration in this area ranges from $3.62$ to $8.60\,\text{ppb}$. Region T2 provides a moderate-concentration baseline, offering an ideal contrast to verify the adaptive exploration and tracking capabilities of the rover.
\end{itemize}

To emulate realistic rover navigation on the lunar surface, where geometric hazards like craters and steep slopes restrict motion, we leverage the corresponding elevation map to extract traversability constraints. Fig.~\ref{fig:scenes} depicts the synthesized helium-3 ground truth alongside the elevation map.

\subsection{Experimental Results and Analysis}

This section evaluates the performance of the active exploration framework from two primary perspectives: the mapping precision of the helium-3 distribution and the traversal efficiency of the planned trajectory.

As there are no established baselines for this exact mission scenario, our experimental evaluation compares the proposed $\text{RRT}^*$ framework against a representative baseline that substitutes the core planner with standard RRT~\cite{lavalle2001randomized}. To evaluate performance, we employ the Root Mean Square Error (RMSE) to measure the helium-3 mapping accuracy, and use the rover’s trajectory length as the metric for exploration efficiency.

\subsubsection{Evaluation of Mapping Precision}

As visualized in the mapping performance curves in Fig.~\ref{fig:RMSE}. Compared to the baseline RRT, which generates longer trajectories that distribute the $N=5$ sampling points across a wider arc, the proposed $\text{RRT}^*$ yields shorter paths where the discrete sampling locations are positioned within a more compact spatial range. 
This structural outcome stems from the decoupling of the information metric from the trajectory geometric optimization, rather than any algorithmic deficiency; thus, it remains fully consistent with theoretical expectations within existing active sensing frameworks. 

The distinct characteristics of regions T1 and T2 in terms of concentration ranges, spatial gradients, and topographic trafficability provide critical context for explaining the subtle performance variations of the planner, such as the opposing trends in RMSE. Consequently, in practical lunar mission planning, the selection of planners and sampling strategies must be tailored to the compositional structure and terrain complexity of the target landing zone, rather than relying on a one-size-fits-all parameter configuration.

\begin{figure}[htbp]
    \centering
    \includegraphics[width=0.5\textwidth]{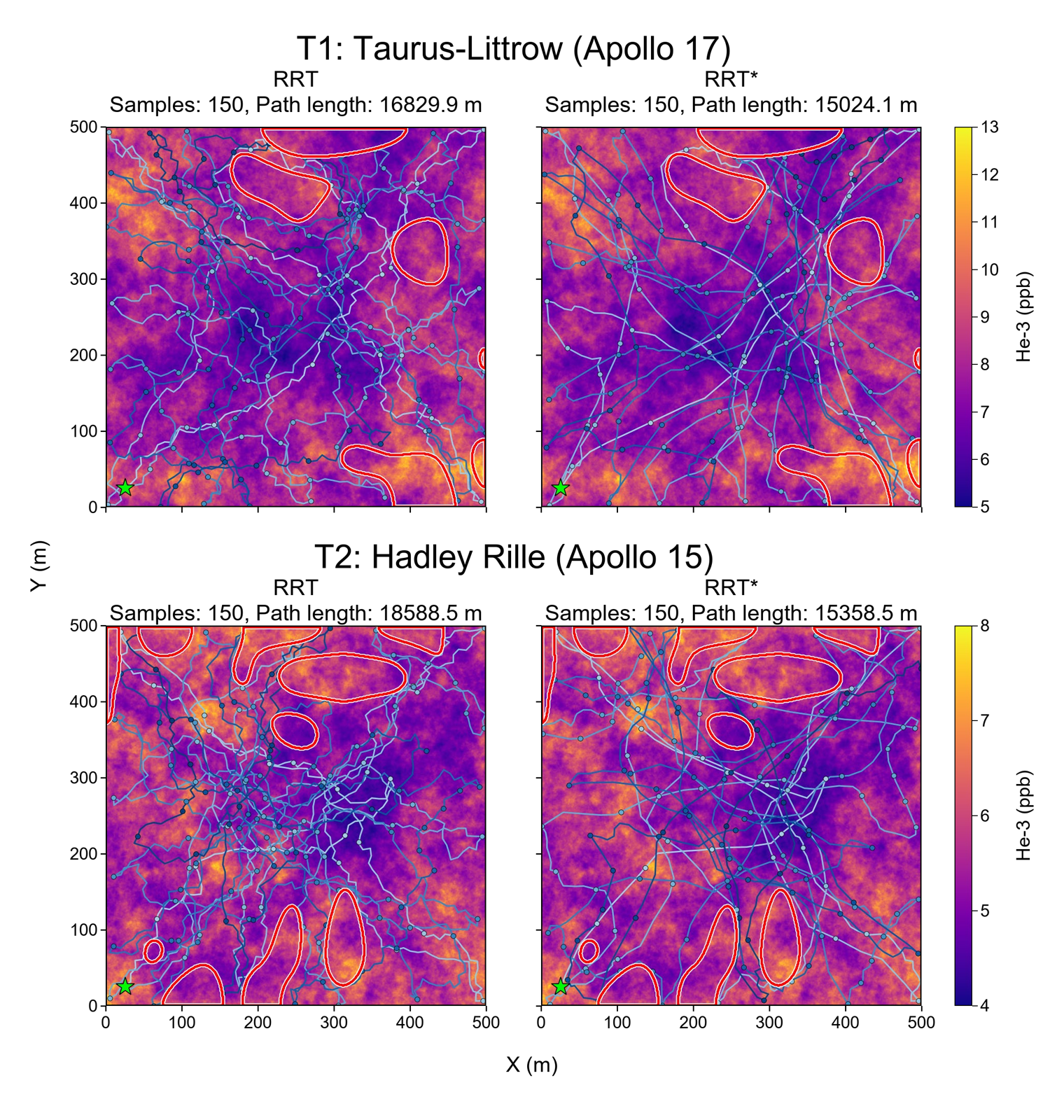} 
    \vspace{-15pt}
    \caption{Path planning results for the two scenarios using RRT* (ours) and RRT (baseline). The circles denote the sampling locations, and the regions circled in red represent non-traversable areas.}
    \label{fig:result} 
\end{figure}

\subsubsection{Evaluation of Trajectory Efficiency}
In terms of path length evaluation, as clearly demonstrated by the trajectory plots in Fig.~\ref{fig:result}, $\text{RRT}^*$ exhibits a robust advantage in traversal efficiency over the standard RRT. Across two independent lunar candidate regions, denoted as T1 and T2, $\text{RRT}^*$ consistently reduces the cumulative trajectory length of the rover. Fig.~\ref{fig:result} illustrates the overall performance evaluation. Specifically, the proposed method achieves a $10.7\%$ reduction in the T1 region and a $17.4\%$ reduction in the T2 region.

This performance gain manifests itself as fewer local deviations and more prolonged straight-line segments. This represents a highly reproducible and stable effect brought by the tree-rewire mechanism, which successfully eliminates redundant kinematic maneuvers during the active information gathering process.

\section{CONCLUSIONS}
This paper concludes by addressing the strategic challenge of lunar helium-3 characterization through a mobile robotic lens. The core of our contribution is the formulation of regolith sampling and drilling-based prospecting within an active exploration and information gathering framework. Utilizing a reliable generative model to simulate non-stationary helium-3 spatial distributions, we introduced He3-Seeker, a systematic robotic informative path planning architecture. The performance of He3-Seeker has been rigorously benchmarked against baselines in simulated hazardous lunar scenarios, demonstrating its advanced capability in optimizing data-gathering efficiency while minimizing traversal costs.

Future work will focus on replicating simulated helium-3 distributions within terrestrial analogues and deploying real robotic platforms for extensive field testing. Ultimately, this paradigm aims to pave the way for autonomous lunar mining and resource extraction technologies.

\bibliographystyle{IEEEtran}
\bibliography{arxiv}

@inproceedings{chen2022ak,
  title={AK: Attentive Kernel for Information Gathering},
  author={Chen, Weizhe and Khardon, Roni and Liu, Lantao},
  booktitle={Robotics: Science and Systems (RSS)},
  year={2022}
}

@inproceedings{smith2020artemis,
  title={The artemis program: An overview of nasa's activities to return humans to the moon},
  author={Smith, Marshall and Craig, Douglas and Herrmann, Nicole and Mahoney, Erin and Krezel, Jonathan and McIntyre, Nate and Goodliff, Kandyce},
  booktitle={IEEE Aerospace Conference},
  pages={1--10},
  year={2020},
}

@article{qiao2026multi,
  title={Multi-Agent Off-World Exploration for Sparse Evidence Discovery via Gaussian Belief Mapping and Dual-Domain Coverage},
  author={Qiao, Zhuoran and Hu, Tianxin and Nguyen, Thien-Minh and Yuan, Shenghai},
  journal={arXiv preprint arXiv:2603.07650},
  year={2026}
}

@inproceedings{booth2023informative,
  title={Informative path planning for scalar dynamic reconstruction using coregionalized Gaussian processes and a spatiotemporal kernel},
  author={Booth, Lorenzo and Carpin, Stefano},
  booktitle={IEEE/RSJ International Conference on Intelligent Robots and Systems (IROS)},
  pages={8112--8119},
  year={2023},
}

@article{li2019china,
  title={China’s present and future lunar exploration program},
  author={Li, Chunlai and Wang, Chi and Wei, Yong and Lin, Yangting},
  journal={Science},
  volume={365},
  number={6450},
  pages={238--239},
  year={2019},
  publisher={American Association for the Advancement of Science}
}

@article{viseras2019robotic,
  title={Robotic active information gathering for spatial field reconstruction with rapidly-exploring random trees and online learning of Gaussian processes},
  author={Viseras, Alberto and Shutin, Dmitriy and Merino, Luis},
  journal={Sensors},
  volume={19},
  number={5},
  pages={1016},
  year={2019},
  publisher={MDPI}
}

@inproceedings{chen2024poam,
  title={POAM: probabilistic online attentive mapping for efficient robotic information gathering},
  author={Chen, Weizhe and Liu, Lantao and Khardon, Roni},
  booktitle={Robotics: Science and Systems (RSS)},
  year={2024}
}

@inproceedings{smith2022viper,
  title={The VIPER Mission, a resource-mapping mission on another celestial body},
  author={Smith, K Ennico and Colaprete, A and Lim, DSS and Andrews, D},
  booktitle={SRR XXII MEETING Colorado School of Mines},
  year={2022}
}

@article{elphic2015simulated,
  title={Simulated real-time lunar volatiles prospecting with a rover-borne neutron spectrometer},
  author={Elphic, Richard C and Heldmann, Jennifer L and Marinova, Margarita M and Colaprete, Anthony and Fritzler, Erin L and McMurray, Robert E and Morse, Stephanie and Roush, Ted L and Stoker, Carol R and Deans, Matthew C and others},
  journal={Advances in Space Research},
  volume={55},
  number={10},
  pages={2438--2450},
  year={2015},
  publisher={Elsevier}
}

@article{geiss2004apollo,
  title={The Apollo SWC experiment: results, conclusions, consequences},
  author={Geiss, J and B{\"u}hler, F and Cerutti, H and Eberhardt, P and Filleux, CH and Meister, J and Signer, P},
  journal={Space Science Reviews},
  volume={110},
  number={3},
  pages={307--335},
  year={2004},
  publisher={Springer}
}

@article{karaman2011sampling,
  title={Sampling-based algorithms for optimal motion planning},
  author={Karaman, Sertac and Frazzoli, Emilio},
  journal={The International Journal of Robotics Research},
  volume={30},
  number={7},
  pages={846--894},
  year={2011},
  publisher={Sage Publications Sage UK: London, England}
}

@article{lavalle2001randomized,
  title={Randomized kinodynamic planning},
  author={LaValle, Steven M and Kuffner Jr, James J},
  journal={The International Journal of Robotics Research},
  volume={20},
  number={5},
  pages={378--400},
  year={2001},
  publisher={SAGE Publications}
}

@article{lucey1998mapping,
  title={Mapping the FeO and TiO2 content of the lunar surface with multispectral imagery},
  author={Lucey, Paul G and Blewett, David T and Hawke, B Ray},
  journal={Journal of Geophysical Research: Planets},
  volume={103},
  number={E2},
  pages={3679--3699},
  year={1998},
  publisher={Wiley Online Library}
}

@article{li2026mining,
  title={Mining beyond Earth with Space Robots: Exploration, Sampling, and Extraction},
  author={Li, Dong and Nie, Dujun and Zhang, Xiaotong and Wang, Ruilin and Li, Yuchen and Ge, Chang and Xiong, Chao and Di, Kaichang and N{\"u}chter, Andreas and Kov{\'a}cs, Levente and others},
  journal={arXiv preprint arXiv:2608.21358},
  year={2026}
}

@article{lemelin2019compositions,
  title={The compositions of the lunar crust and upper mantle: Spectral analysis of the inner rings of lunar impact basins},
  author={Lemelin, Myriam and Lucey, Paul G and Miljkovi{\'c}, Katarina and Gaddis, Lisa R and Hare, Trent and Ohtake, Makiko},
  journal={Planetary and Space Science},
  volume={165},
  pages={230--243},
  year={2019},
  publisher={Elsevier}
}

@article{fang2014lunar,
  title={Lunar Penetrating Radar onboard the Chang'e-3 mission},
  author={Fang, Guang-You and Zhou, Bin and Ji, Yi-Cai and Zhang, Qun-Ying and Shen, Shao-Xiang and Li, Yu-Xi and Guan, Hong-Fei and Tang, Chuan-Jun and Gao, Yun-Ze and Lu, Wei and others},
  journal={Research in Astronomy and Astrophysics},
  volume={14},
  number={12},
  pages={1607--1622},
  year={2014}
}

@article{lawrence1998global,
  title={Global elemental maps of the Moon: The Lunar Prospector gamma-ray spectrometer},
  author={Lawrence, DJ and Feldman, WC and Barraclough, BL and Binder, AB and Elphic, RC and Maurice, S and Thomsen, DR},
  journal={Science},
  volume={281},
  number={5382},
  pages={1484--1489},
  year={1998},
  publisher={American Association for the Advancement of Science}
}

@article{zhang2025he,
  title={He, Ne, and Ar isotope systematics in Chang’e-5 plagioclase reveal diffusive loss and reirradiation processes},
  author={Zhang, Xuhang and Su, Fei and Avice, Guillaume and Bekaert, David V and Obase, Tomoya and Otsuki, Yuta and Stuart, Finlay M and Zhang, Yingnan and Nie, Jiayan and Li, Xiaoguang and others},
  journal={Earth and Planetary Science Letters},
  volume={671},
  pages={119666},
  year={2025},
  publisher={Elsevier}
}

@article{jolliff2000major,
  title={Major lunar crustal terranes: Surface expressions and crust-mantle origins},
  author={Jolliff, Bradley L and Gillis, Jeffrey J and Haskin, Larry A and Korotev, Randy L and Wieczorek, Mark A},
  journal={Journal of Geophysical Research: Planets},
  volume={105},
  number={E2},
  pages={4197--4216},
  year={2000},
  publisher={Wiley Online Library}
}

@article{fa2007quantitative,
  title={Quantitative estimation of helium-3 spatial distribution in the lunar regolith layer},
  author={Fa, Wenzhe and Jin, Ya-Qiu},
  journal={Icarus},
  volume={190},
  number={1},
  pages={15--23},
  year={2007},
  publisher={Elsevier}
}

@article{ding2023moon,
  title={Moon-based ground penetrating radar derivation of the Helium-3 reservoir in the regolith at the Chang'E-3 landing site},
  author={Ding, Chunyu and Li, Qingquan and Xu, Jiangwan and Lei, Zhonghan and Li, Jiawei and Su, Yan and Huang, Shaopeng},
  journal={IEEE Journal of Selected Topics in Applied Earth Observations and Remote Sensing},
  volume={16},
  pages={2764--2776},
  year={2023},
  publisher={IEEE}
}

@article{simko2014lunar,
  title={Lunar helium-3 fuel for nuclear fusion: Technology, economics, and resources},
  author={Simko, Thomas and Gray, Matthew},
  journal={World Future Review},
  volume={6},
  number={2},
  pages={158--171},
  year={2014},
  publisher={SAGE Publications Sage CA: Los Angeles, CA}
}

@article{lucey2000lunar,
  title={Lunar iron and titanium abundance algorithms based on final processing of Clementine ultraviolet-visible images},
  author={Lucey, Paul G and Blewett, David T and Jolliff, Bradley L},
  journal={Journal of Geophysical Research: Planets},
  volume={105},
  number={E8},
  pages={20297--20305},
  year={2000},
  publisher={Wiley Online Library}
}

\end{document}